\documentclass[letterpaper, 10 pt, conference]{ieeeconf}

\IEEEoverridecommandlockouts

\usepackage{graphicx}
\usepackage{multirow}
\usepackage{multicol}
\usepackage{url}
\usepackage{amsmath}
\usepackage{amsfonts}
\usepackage{subcaption}
\usepackage{tabularx}
\usepackage{hyperref}
\usepackage{amsmath}
\allowdisplaybreaks

\usepackage{tikz}
\usepackage{lipsum}
\usepackage{hyperref}
\newcommand\preprinttext{%
    \scriptsize This is a preprint - Published in IEEE/RSJ International Conference on Intelligent Robots and Systems (IROS) 2025 - DOI: \href{https://doi.org/10.1109/IROS60139.2025.11246746}{10.1109/IROS60139.2025.11246746}
}
\newcommand\copyrighttext{
    \scriptsize \textcopyright 2025 IEEE. Personal use of this material is permitted. Permission from IEEE must be obtained for all other uses, in any current or future media, including reprinting/republishing this material for advertising or promotional purposes, creating new collective works, for resale or redistribution to servers or lists, or reuse of any copyrighted component of this work in other works.
}
\newcommand\noticeblock{
    \begin{tikzpicture}[remember picture,overlay]
    \node[anchor=north,yshift=-30pt] at (current page.north) {\preprinttext};
    \node[anchor=south,yshift=20pt] at (current page.south) {
        \parbox{\dimexpr\textwidth-\fboxsep-\fboxrule\relax}{\copyrighttext}
    };
    \end{tikzpicture}
}

\title{\LARGE \bf
    NaviFormer: A Deep Reinforcement Learning Transformer-like Model to Holistically Solve the Navigation Problem
}

\author{
    Daniel Fuertes$^{1}$,
    Andrea Cavallaro$^{2}$,
    Carlos R. del-Blanco$^{1}$,
    Fernando Jaureguizar$^{1}$ and
    Narciso García$^{1}$
    \thanks{
        Daniel Fuertes wishes to thank the “Ayudas dirigidas al personal investigador en formación predoctoral para realizar una estancia de investigación internacional igual o superior a tres meses”, funded by Programa Propio I+D+i 2023 from Universidad Politécnica de Madrid. This work has been partially supported by project PID2023-148922OA-I00 (EEVOCATIONS) funded by MCIU/AEI/10.13039/501100011033 of the Spanish Government, and by project TEC-2024/COM-322 (IDEALCV-CM) funded by Comunidad de Madrid.
    }
    \thanks{
        $^{1}$Daniel Fuertes, Carlos R. del-Blanco, Fernando Jaureguizar, and Narciso García are with Grupo de Tratamiento de Imágenes, Information Processing and Telecommunications Center, ETSI Telecomunicación, Universidad Politécnica de Madrid, 28040 Madrid, Spain.
        {\tt\small \{d.fcoiras, carlosrob.delblanco, fernando.jaureguizar, narciso.garcia\}@upm.es}
    }
    \thanks{
        $^{2}$Andrea Cavallaro is with Idiap Research Institute, Martigny, Switzerland.
        {\tt\small a.cavallaro@idiap.ch}
    }
}

\begin{document}

\maketitle
\thispagestyle{empty}
\pagestyle{empty}
\noticeblock

\begin{abstract}
Path planning is usually solved by addressing either the (high-level) route planning problem (waypoint sequencing to achieve the final goal) or the (low-level) path planning problem (trajectory prediction between two waypoints avoiding collisions). However, real-world problems usually require simultaneous solutions to the route and path planning subproblems with a holistic and efficient approach. In this paper, we introduce NaviFormer, a deep reinforcement learning model based on a Transformer architecture that solves the global navigation problem by predicting both high-level routes and low-level trajectories. To evaluate NaviFormer, several experiments have been conducted, including comparisons with other algorithms. Results show competitive accuracy from NaviFormer since it can understand the constraints and difficulties of each subproblem and act consequently to improve performance. Moreover, its superior computation speed proves its suitability for real-time missions.
\end{abstract}

\section{Introduction}
\label{sec:intro}

Planning tours for agents is a complex task with multiple applications, such as package delivery, transportation, film-making, surveillance, search-and-rescue, exploration, and agriculture. The general idea is to generate routes for a given ground, aerial, or underwater unmanned vehicle that connect a set of waypoints/regions where the agent is expected to perform one or several tasks. Real-world scenarios usually impose constraints such as limited energy (battery or fuel) budget, which may prevent agents from visiting all regions; or static/dynamic obstacles, which may force to plan a new, typically suboptimal solution. These real-world scenarios are usually modeled as two different subproblems: route planning and path planning.

Route planning consists of high-level planning to infer the best sequence of waypoints (route) without considering the exact trajectory between pairs of waypoints. It is frequently modeled as a Vehicle Routing Problem (VRP) \cite{Mor2022}. However, VRPs do not usually consider obstacles and assume that every path is the straight line (Euclidean distance) connecting two waypoints. Instead, path planning \cite{Karur2021} focuses on finding the shortest low-level trajectory (path) defined by a discretized sequence of coordinates. Contrary to route planning, path planning considers obstacles that hinder the travel from a start to a goal waypoint. However, it is not involved in deciding the next node for a high-level objective.

\begin{figure}[t]
\centering
    \begin{subfigure}{.325\columnwidth}
      \centering
      \includegraphics[width=\columnwidth]{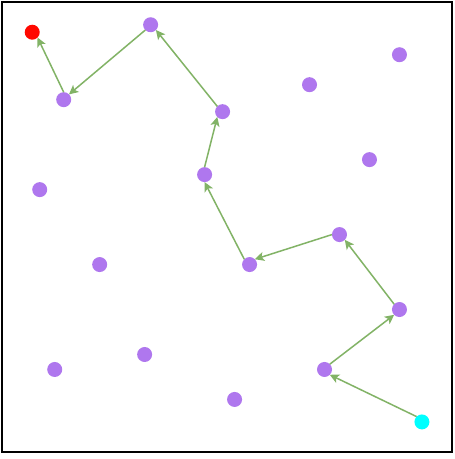}
      \caption{Route planning.}
      \label{fig:route}
    \end{subfigure}
    \begin{subfigure}{.325\columnwidth}
      \centering
      \includegraphics[width=\columnwidth]{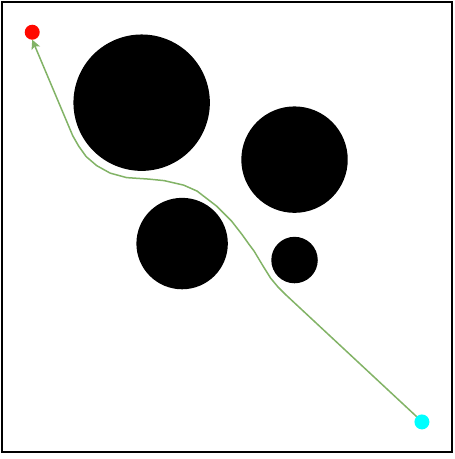}
      \caption{Path planning.}
      \label{fig:path}
    \end{subfigure}
    \begin{subfigure}{.325\columnwidth}
      \centering
      \includegraphics[width=\columnwidth]{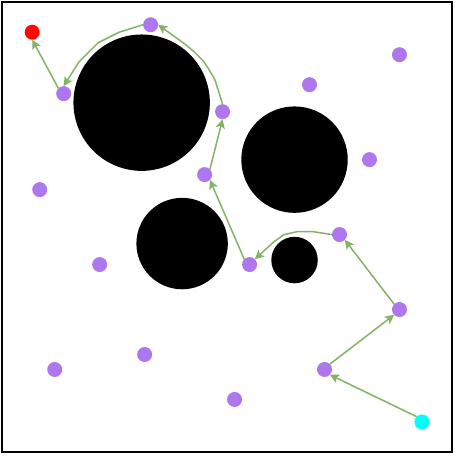}
      \caption{Nav. planning.}
      \label{fig:navigation}
    \end{subfigure}
\caption{Example of: (a) Route planning - maximize number of visited regions (purple circles) from an initial to an end depot (cyan/red circles) within a time limit (similar to OP); (b) Path planning - find the shortest path to a goal region while avoiding obstacles (black circles); and (c) Navigation planning - holistic combination of both subproblems.}
\label{fig:problems}
\end{figure}

In this paper, we propose a new and holistic definition of the problem, called Navigation Orienteering Problem (NOP), based on the Orienteering Problem (OP) \cite{Golden1987}. Previous works usually focus on either path planning \cite{Jin2023,Yonetani2021,Lawson2023} or route planning \cite{Kool2019,Fuertes2022, Ortools2023}, and include the constraints that best fit the target application. On the contrary, we define a combination of both subproblems (see Fig. \ref{fig:problems}) including basic constraints that would fit any application, such as the time limit due to energy consumption (similar to OP); the presence of obstacles, which exist in many real-world scenarios; and the arrival to an end depot (also similar to OP). The NOP introduces a new challenge where the agent must provide collision-free trajectories that visit several waypoints to ultimately achieve a high-level objective.

Our main contribution is the design of a novel navigation system based on deep reinforcement learning (DRL), called NaviFormer, that solves the holistic NOP in real-time. NaviFormer approaches each individual subproblem (route planning and path planning) in a global manner, allowing better understanding of the environment and enhancing the quality of solutions. It is based on a Transformer neural network trained with DRL to solve the NOP. This Transformer-based network is composed of an encoder-decoder architecture capable of encoding graphs of nodes/waypoints, and combining them with information from the 
obstacles of the environment using a special combined attention operation. The resulting embedding, containing global information from the environment, is used to decode the path that maximizes the system performance. Thus, the encoder generates a semantic representation of a scenario by computing a node-graph embedding, while the decoder iteratively predicts, at each time step, the next waypoint to visit and the next direction to follow, allowing the agent to avoid obstacles.

\section{Related works}
\label{sec:sota}

Although there are a few works that propose a direct heuristic solution to the navigation problem, such as \cite{Shi2024} 
, most of them \cite{Lu2023} address the navigation problem as two independent subproblems: route and path planning.


Route planning approaches can be divided into three main categories: linear optimizers, heuristic algorithms, and machine learning. Linear optimizers are computationally expensive methods that seek the optimal solution within a defined set of feasible solutions, subject to linear constraints. Examples include cutting planes, especially for multiple agents \cite{Sundar2022}, and commercial solvers such as OR-Tools \cite{Ortools2023}, or Gurobi \cite{Gurobi2023}. Although accurate, these methods are usually unsuitable for real-time operations.

To reduce computation time, heuristics \cite{Purkayastha2020} and metaheuristics \cite{Rahman2021} are frequently applied. These methods sacrifice some precision in favor of finding approximate solutions within a reduced amount of time. Some approaches include Variable Neighborhood Search (VNS) \cite{Bezerra2023}, or Greedy Randomized Adaptive Search Procedure (GRASP) \cite{Bruglieri2022}. Bioinspired algorithms, such as Genetic Algorithms (GA) \cite{Xiao2022}, Particle Swarm Optimization (PSO) \cite{Xiao2022}, and Ant Colony Optimization (ACO) \cite{Xiao2022}, usually fall into the category of metaheuristics and can also be applied to VRPs. Although (meta) heuristic methods are good alternatives to improve speed, they still tend to not reach real-time performance.

DRL frameworks are particularly interesting for routing problems, as they employ neural networks that learn from the experience by trying actions and receiving rewards from the environment. Unlike linear optimization and heuristic algorithms, which refine solutions through iterative processes within the feasible solution space, DRL methods first encode data from the environment and then yield predictions that maximize the received reward. Current works use different architectures, such as Convolutional Neural Networks (CNN) \cite{Jung2024} or Graph Neural Networks (GNN) \cite{Guo2024}, to encode the graph of nodes that represent routing problems. For the decoding step, responsible for sequentially predicting routes, some works have proposed to combine Attention models and Recurrent Neural Networks (RNN) \cite{Yang2021} to obtain Pointer Networks (PN) \cite{Gama2021}. However, recent Transformer networks \cite{Kool2019,Fuertes2022} have outperformed RNN-based methods due to their faster parallel-like data processing, and the robustness of multi-head attention encoding, specially for long data sequences.


Path planning, works in this area can also be classified into the same categories. However, real-time performance becomes more critical due to obstacle avoidance constraints. Thus, even if linear optimizers are used in some cases, like cutting planes \cite{Lam2022}, they are not popular. 

Graph search algorithms, such as A* \cite{Mandloi2021}, D* \cite{Ravankar2017}, and D* Lite \cite{Jin2023}, exploit heuristic and sampling techniques to accelerate convergence. A* uses admissible heuristics to guide Dijkstra's graph search and find optimal solutions. D* and D* Lite improve A* speed and allow dynamic obstacle handling. Other sampling strategies include Probabilistic Roadmaps (PRM) \cite{Lawson2023}, which constructs a roadmap based on collision-free path probabilities, and Artificial Potential Fields (APF) \cite{Pan2022}, which uses attractive and repulsive forces for navigation. Furthermore, bioinspired learning methods, such as GA \cite{Zhang2023}, ACO \cite{Wu2023}, and PSO \cite{Yu2022}, also use (meta) heuristics for path planning.

Some of the mentioned approaches perform very fast but yield approximate solutions with limited performance. Instead, machine learning methods, especially those focused on DRL, have the potential to learn representations of the environment to find better solutions. They often incorporate a CNN to encode binary global maps (representing the whole scenario) and make predictions through a set of dense layers \cite{Liu2020}. Some methods \cite{Bartolomei2022} extend this CNN-like structure by incorporating depth maps to the environment analysis. Alternatively, Transformers have replaced CNN encoders with more powerful Vision Transformers (ViT) \cite{LinChen2023}. In contrast, our approach proposes to dynamically encode reduced local maps representing the agent's surroundings by a lightweight CNN, enabling fast and efficient predictions. Moreover, it jointly addresses the interrelated problems of route and path planning, reaching better performance than decomposing the problem into two independent subproblems.
\section{Problem formulation}
\label{sec:problem}

Consider a set of nodes $G = \{0, ..., n + 1\}$ representing the visitable regions, where the nodes $0$ and $n + 1$ are the start and end depots. An agent is expected to visit the nodes and reach the end depot within a time limit $T$. The NOP seeks routes that are rewarded for visiting regions of $G$ with a set of rewards $R = \{ r_0, ..., r_{n+1} \}$ while avoiding a set of $b$ obstacles $O = \{ o_i |i = 0, ..., b\}$ represented as circles, where the $i^{th}$ obstacle is parameterized by its center and radius as $o_i=(x^{obs}, y^{obs}, \ r^{obs})$. The path followed is discretized with a step length of $t_s$, such that the total number of steps allowed is $L = \left\lfloor \frac{T}{t_s} \right\rfloor$. The objective of the NOP is to maximize the constrained function of Eq. \ref{eq:goalFunc}.

\vspace{-3mm}
\begin{flalign}
    \max_{\Phi} \ \ \
        \sum_{i = 0}^{n} &
            \sum_{j = 1}^{n+1}
                r_j \Phi_{ij}, \
        \Phi_{ij} =
            \begin{cases}
                1, & i \ \text{is right after} \ j \\
                0, & \text{otherwise}
            \end{cases}
    & \label{eq:goalFunc}
\end{flalign}

  \begin{flalign}
    \label{eq:constraint1}
        \textrm{s.t.} \quad
            \sum_{j = 1}^{n + 1}
                \Phi_{0j} = 1
    \qquad \qquad \qquad \qquad \ \ \ \\
    \label{eq:constraint2}
            \sum_{i = 0}^{n}
                \Phi_{i(n + 1)} = 1
    \qquad \qquad \qquad \quad \ \\
    \label{eq:constraint3}
        \sum_{i = 1}^{n}
            \Phi_{ij} \in \{0, 1\};
            \quad j \in \{1, ..., n\}
    \ \\
    \label{eq:constraint4}
        \sum_{j = 1}^{n}
            \Phi_{ij} \in \{0, 1\};
            \quad i \in \{1, ..., n\}
    \ \\
    \label{eq:constraint5}
        \sum_{i = 0}^{n+1}
            \Phi_{ii} = 0
    \qquad \qquad \qquad \quad \quad \ \ \ \\
    \label{eq:constraint6}
        L_{0(n+1)} \leq
            \sum_{i=0}^{n}
                \sum_{j=0}^{n+1}
                    L_{ij} \Phi_{ij} \leq L
    \quad \ \\
    \label{eq:constraint7}
        L_{ij} =
            \min_{\nu_{i, j} \in \mathcal{V}_{i, j}}
                d(\nu_{i, j}); \
                i, j \in G
    \quad \ \\
    \label{eq:constraint8}
        u_{i} - u_{j} + n\Phi_{ij} \leq n - 1; \
        i,j \in G
  \end{flalign}

Constraints of Eqs. \ref{eq:constraint1} and \ref{eq:constraint2} force the agent to start and finish the paths at regions $0$ and $n+1$, respectively. Eqs. \ref{eq:constraint3} and \ref{eq:constraint4} ensure continuous routes and prevent revisiting nodes, while Eq. \ref{eq:constraint5} forbids immediate node revisits. Eq. \ref{eq:constraint6} limits the distance/time budget $T$ (we assume that the agent moves at a constant speed, meaning that the time limit can be converted to a distance limit), which is discretized as $L$, and imposes that this budget allows at least to travel from node $0$ to node $n+1$ ($L_{0(n+1)}$). Restriction of Eq. \ref{eq:constraint7} minimizes path length between nodes ($L_{i,j}$), where $d(\nu_{i, j})$ is the length of the path $\nu_{i, j} = \{ (x,y) \ | \ x,y \ \in \ [0,1] \ \wedge \ x,y \ \notin \ O \}$ connecting $i$ and $j$, and $\mathcal{V}_{i, j}$ is the set of all paths connecting $i$ and $j$ without colliding any obstacle from $O$. Finally, subtours \cite{Vansteenwegen2011} are avoided thanks to Eq. \ref{eq:constraint8}, where $u_{i}, u_{j} \in \{1, ..., n\}$ are the positional order of $i$ and $j$ on the path.
\section{NaviFormer Neural Network}
\label{sec:system}

\begin{figure*}[t]
    \centerline{\includegraphics[width=0.8\textwidth]{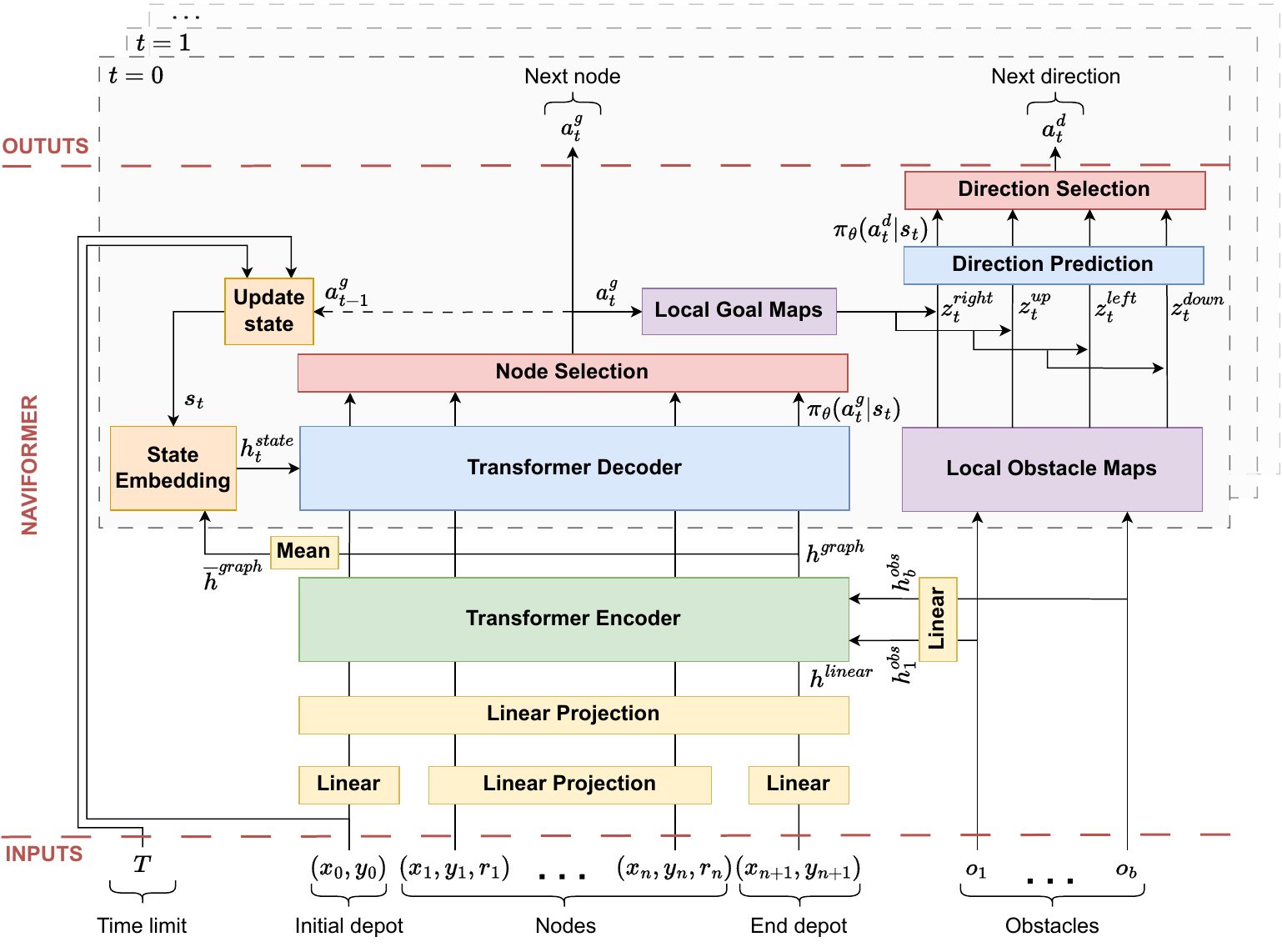}}
    \caption{
    NaviFormer's architecture: a modified Transformer encoder combines simple linear representations of nodes ($h^{lin}$) and obstacles ($h^{obs}$) into a more complex graph embedding ($h^{graph}$). Later, a Transformer decoder utilizes $h^{graph}$ and $h^{state}$ (encoding of the agent’s state $s_t$ from the state embedding module) to iteratively predict a policy $\pi_{\theta} (a_t^g | s_t)$, from which the next goal node $a_t^g$ can be sampled. Similarly, the direction prediction module infers the policy $\pi_{\theta} (a_t^d | s_t)$ from $a_t^g$ and $o_1,...,o_b$ to obtain the next direction to follow $a_t^d$.
    }
    \label{fig:network}
\end{figure*}

\begin{figure*}[t]
\centering
    \begin{subfigure}{0.4\textwidth}
      \centering
      \includegraphics[width=\textwidth]{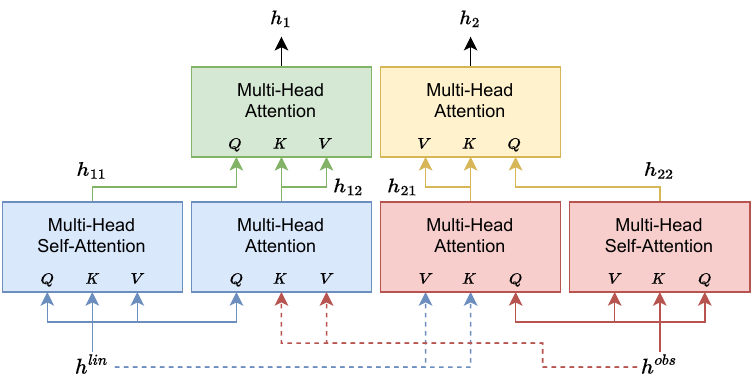}
      \caption{Combined multi-head attention module.}
      \label{fig:combined_mha}
      \centering
      \includegraphics[width=\textwidth]{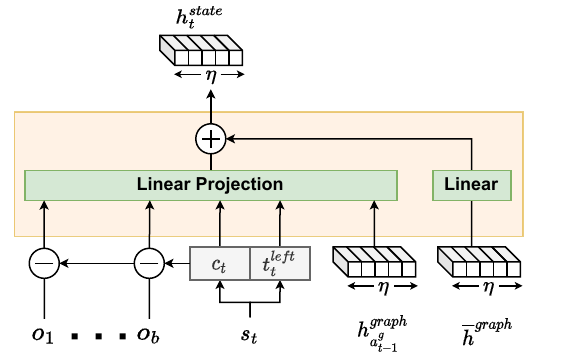}
      \caption{State embedding module.}
      \label{fig:state}
    \end{subfigure}
    \begin{subfigure}{0.57\textwidth}
      \centering
      \includegraphics[width=\textwidth]{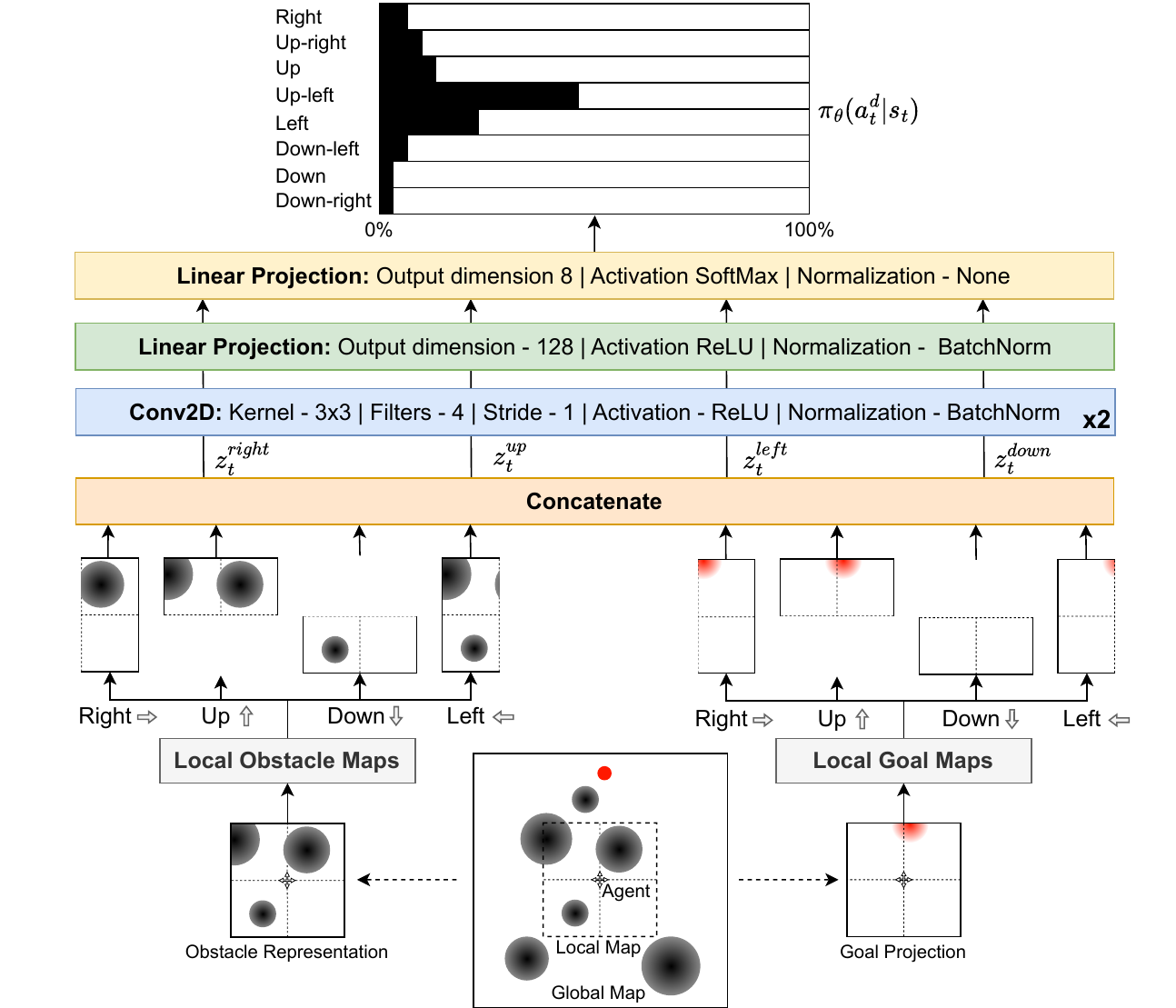}
      \caption{Direction prediction layers.}
      \label{fig:direction}
    \end{subfigure}
\caption{Novel NaviFormer modules: (a) the combined multi-head attention operation to merge node and obstacle information, (b) the state embedding module to encode the situation of the agent at each time step $t$, and (c) the direction prediction layers to predict the policy $\pi_{\theta}(a^d_t|s_t)$ from local maps based on the obstacles and the next goal $a^g_t$.}
\label{fig:modules}
\end{figure*}

Since the NOP imposes to find a path that maximizes the visited regions, we let our network predict action $a^d_t \in D = \{ \frac{\pi}{4}k \ | \ k = 0, ..., 7 \}$ indicating the direction the agent should follow at each time step $t \in \{1, ..., L\}$. These directions should construct a trajectory that could solve the problem. However, we wanted to improve the network insight about the scenario by letting it also predict each next goal to visit $a^g_t \in G$. Note that actions $a^d_t$ and $a^g_t$ solve Eqs. \ref{eq:goalFunc} and \ref{eq:constraint7}.

Therefore, NaviFormer, depicted in Fig. \ref{fig:network}, is based on an encoder-decoder Transformer architecture that first encodes the regions to travel to, characterized by their position ($x_i, y_i; i \in G$) and reward ($r_i$), and the obstacles to avoid, both as a graph of nodes ($h^{graph}$). This embedding is a projection into a $\eta$-dimensional feature space to extract relevant and discriminating information. Then, it uses that information to decode and predict a policy $\pi_{\theta}(a^g_t, a^d_t|s_t)$ that represents the probability distribution of possible actions $a^g_t, a^d_t$ on each state $s_t$. The prediction of this policy also depends on two additional modules: a state embedding ($h^{state}_t$), with information about the spatial position of the agent and the elapsed time, combined with $h^{graph}$ to find the best next goal ($\pi_{\theta}(a^g_t|s_t)$); and a direction predictor, where a set of local maps allows finding the best direction ($\pi_{\theta}(a^d_t|s_t)$). In this manner, our approach can infer efficient solutions restricted by the problem constraints from Sec. \ref{sec:problem}.


\textbf{Encoder} NaviFormer encoder, based on standard route planning Transformer encoders like \cite{Kool2019,Fuertes2022}, receives individual linear projections of the input nodes $h^{lin}$ and obstacles $h^{obs}$ with dimension $\eta=128$, and generates a combined graph embedding $h^{graph}$ by learning some attention scores $S$ that promote those node connections that improve the expected reward in the long-term. Contrary to standard encoders that combine pairs of input data, NaviFormer encoder considers a three-way relationship to find the affinity between every pair of nodes with respect to each obstacle. For that purpose, the combined multi-head attention strategy of Fig. \ref{fig:combined_mha} is proposed to substitute standard self-attention mechanisms. It takes the feature vectors $h^{lin}$ and $h^{obs}$, and applies a self-attention operation to find the affinity between nodes from the same set, and an attention operation between the two embeddings $h^{lin}$ and $h^{obs}$ to find the crossed-affinity. The output of both operations ($h_{11}$ and $h_{12}$ from $h^{lin}$, and $h_{21}$ and $h_{22}$ from $h^{obs}$) are merged through another pair of attention layers to obtain the final combined embeddings $h_1$ and $h_2$ as follows.

\vspace{-5mm}
\begin{flalign} 
    & h_{11} = \text{A}(h^{lin}, h^{lin}, h^{lin}) &
    & h_{21} = \text{A}(h^{obs}, h^{lin}, h^{lin}) & \nonumber \\
    & h_{12} = \text{A}(h^{lin}, h^{obs}, h^{obs}) &
    & h_{22} = \text{A}(h^{obs}, h^{obs}, h^{obs}) & \nonumber\\
    & h_{1} = \text{A}(h_{11}, h_{12}, h_{12}) &
    & h_{2} = \text{A}(h_{22}, h_{21}, h_{21}) &
\end{flalign}

where $\text{A}(h^Q, h^K, h^V)$ is the multi-head attention operation. Notice the difference between self-attention (same embedding for query, key, and value), and (crossed) attention (different embeddings). The encoding block of the Transformer is stacked $N$ times to obtain a deeper model. For the last block, $h_{2}$ is not necessary since the resulting $h^{graph}$ is just inferred from $h_{1}$.


\textbf{State Embedding} In addition to the scenario encoding, the agent's state $s_t$ is also encoded by the state embedding module (see Fig. \ref{fig:state}). It includes the agent's position $c_t$ and time $t_t^{left}$, the distance to the obstacles $o_1, ..., o_b$, and the agent's provenance (included in the graph embedding of the last node $h^{graph}_{a^g_{t-1}}$). The final state embedding ($h^{state}_t$) is obtained by adding the linear projection of the aforementioned elements and the linear projection of the averaged graph embedding across all nodes $\overline{h}^{graph}$ that provides some context about the agent's location on the graph.


\textbf{Decoder} Unlike the encoder, which returns a unique static $h^{graph}$, the decoder predicts actions at every time step from the scenario ($h^{graph}$) and the agent's state ($h^{state}_t$) autoregressively. The first module (masked multi-head attention) combines both inputs with an attention layer that includes a mask $M$ to satisfy specific constraints of Section \ref{sec:problem}.

\vspace{-3mm}
\begin{flalign}\label{eq:mask}
     S(h^{state}_t, \ h^{graph}) & = \text{SoftMax} \left(
            M \frac{
                Q(h^{state}_t) K(h^{graph})^T
            }{\sqrt{\eta}}
        \right), & \nonumber \\
    M_i & =
        \begin{cases}
            -\infty & \text{if} \ i \ \text{visited} \\
            1 & \text{otherwise}
        \end{cases}, \forall i \in G &
\end{flalign}

This mask ensures that no region is visited multiple times. The last module is a masked single-head attention that uses a tanh activation function and a unique head to predict a multinomial probability distribution $\pi(a^g_t|s_t)$, from which the next node $a^g_t$ is sampled.


\textbf{Direction Prediction} In addition to $\pi(a^g_t|s_t)$, NaviFormer estimates $\pi_{\theta}(a^d_t|s_t)$ to obtain a prediction of the agent's direction. For this purpose, local maps (see Fig. \ref{fig:direction}) centered at the agent's position are generated. The local obstacle map and the next selected node map (projected to appear on the maps) are divided into four sections representing the agent's north ($z^{up}_t$), east ($z^{right}_t$), south ($z^{down}_t$), and west ($z^{left}_t$). These sections provide knowledge about the scenario to predict the best direction to reach the next node while avoiding obstacles. Finally, the policy $\pi_{\theta}(a^d_t|s_t)$ is obtained feeding the local maps to a lightweight CNN, composed of a couple of convolutional and dense layers.


\textbf{Training Strategy} NaviFormer is trained with DRL by simulating episodes of different problem instances $\alpha$ and collecting, for each $\alpha$, the following reward values:

\vspace{-3mm}
\begin{flalign}\label{eq:rewards}
    r^{\pi_{\theta}}(\alpha) & =
        \sum_{a^g \in A^g}
            \gamma\frac{r_{a^g}}{n/2}
            - \beta \sum_{a^d \in A^d_{a^g}} d(c_t, a^g)
            + \xi, & \\
    \xi & = \begin{cases}
        +20 & \text{if the episode is successful} \\
        -10 & \text{otherwise}
    \end{cases} & \nonumber
\end{flalign}

where $r^{\pi_{\theta}}(\alpha)$ is the reward received after following $\pi_{\theta}$ in $\alpha$, $A^g$ is the set of nodes visited, $A^d_{a^g}$ is the set of directions to reach $a^g$, $d(c_t, a^g)$ is the distance between $c_t$ and $a^g$, $\xi$ is a reward/penalization given if the agent is successful or not (if it reaches the end depot on time without colliding), and $\gamma=10$, $\beta=0.3$ are constant values. Besides, the reward for visiting each region is $r_{a^g} = 1$ if $ a^g \in \{ 1, ..., n \}$ and 0 otherwise. To maximize the reward collection for the NOP (Eq. \ref{eq:goalFunc}), we extend Reinforce to a vanilla Actor-Critic by using a critic value $V^{\pi_\theta}(\alpha)$ as baseline $b(\alpha)$, which reduces the variance of the obtained reward across different $\alpha$ and improves the learning process. The gradient of the resulting loss function for the gradient descent update is defined below.

\vspace{-3mm}
\begin{equation} \label{eq:actor_critic}
    \nabla \mathcal{L}(\theta | \alpha) =
        E_{ \pi_{\theta} (\nu | \alpha)} \left[
            (r^{\pi_{\theta}}(\alpha) - b(\alpha))
            \nabla \log \pi_{\theta} (\nu | \alpha)
        \right]
\end{equation}

where $\nu$ is the set of actions $(a^g, a^d)$ that define a path, and $b(\alpha) = V^{\pi_\theta}(\alpha)$ is a critic baseline predicted from $\overline{h}^{graph}$ by applying two dense layers of hidden size $\eta=128$ connected through a ReLU activation function. Additionally, to encourage the exploration of new actions during training, we sample them from the multinomial distribution $\pi_{\theta}(a_t|s_t)$ so that the best known actions are more likely to be exploited, but other actions could also be explored. During inference, greedy selection is performed instead. Other training details include using an Adam optimizer with learning rate of $10^{-4}$, which achieves convergence after approximately 100 epochs.


\textbf{Constraint Satisfaction} Some of the constraints of the NOP proposed in Sec. \ref{sec:problem} are directly imposed on NaviFormer (hard constraints) while others are learned (soft constraints). Hard constraints include starting the path at node $0$ (Eq. \ref{eq:constraint1}), preventing revisiting nodes (Eqs. \ref{eq:constraint3}-\ref{eq:constraint5}) and avoiding subtours (Eq. \ref{eq:constraint8}). Eq. \ref{eq:constraint1} is inherently satisfied by initializing the agent’s state at the position of the initial depot. Instead, Eqs. \ref{eq:constraint3}-\ref{eq:constraint5}, and \ref{eq:constraint8} are satisfied using mask $M$, described in Eq. \ref{eq:mask}. This mask assigns a zero probability to predicting already-visited nodes by setting the non-normalized logits to $-\infty$. On the other hand, soft constraints are learned using the $\xi$ and $d(c_t, a^g)$ terms from Eq. \ref{eq:rewards} to reward the agent if it successfully reaches the end depot (Eq. \ref{eq:constraint2}) on time (Eq. \ref{eq:constraint6}) and minimizes the length of the collision-free path between each pair of visited nodes (Eq. \ref{eq:constraint7}).

\section{Results}
\label{sec:results}

\begin{figure}[t]
\centering
    \begin{subfigure}{\columnwidth}
      \centering
    \includegraphics[width=0.8\textwidth]{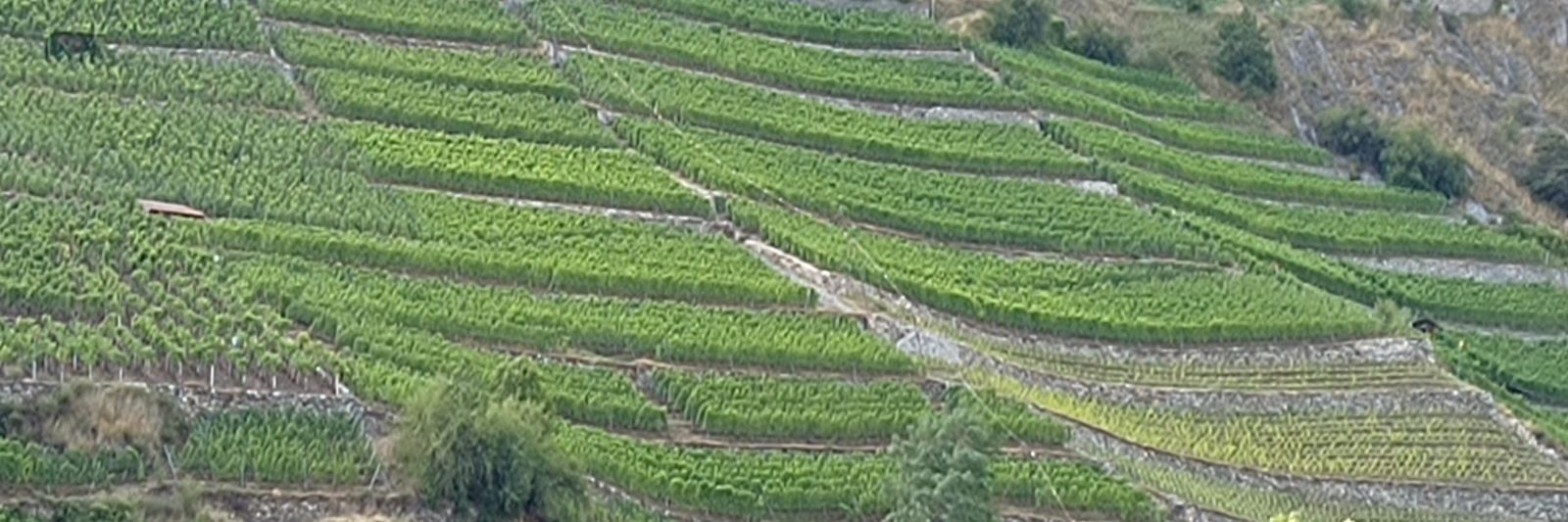}
      \caption{Agriculture scenario.}
      \label{fig:vineyard}
    \end{subfigure}
    \begin{subfigure}{\columnwidth}
      \centering
      \includegraphics[width=0.8\textwidth]{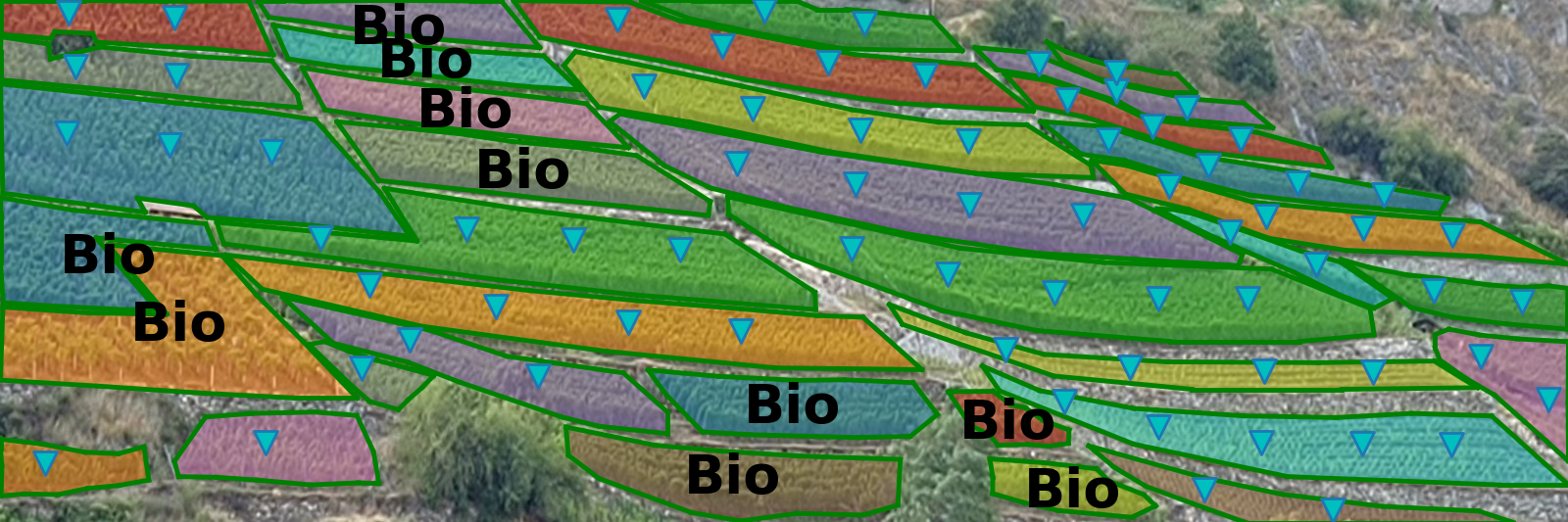}
      \caption{Spraying points of every area.}
      \label{fig:spraying}
    \end{subfigure}
    \begin{subfigure}{\columnwidth}
      \centering
      \includegraphics[width=0.8\textwidth]{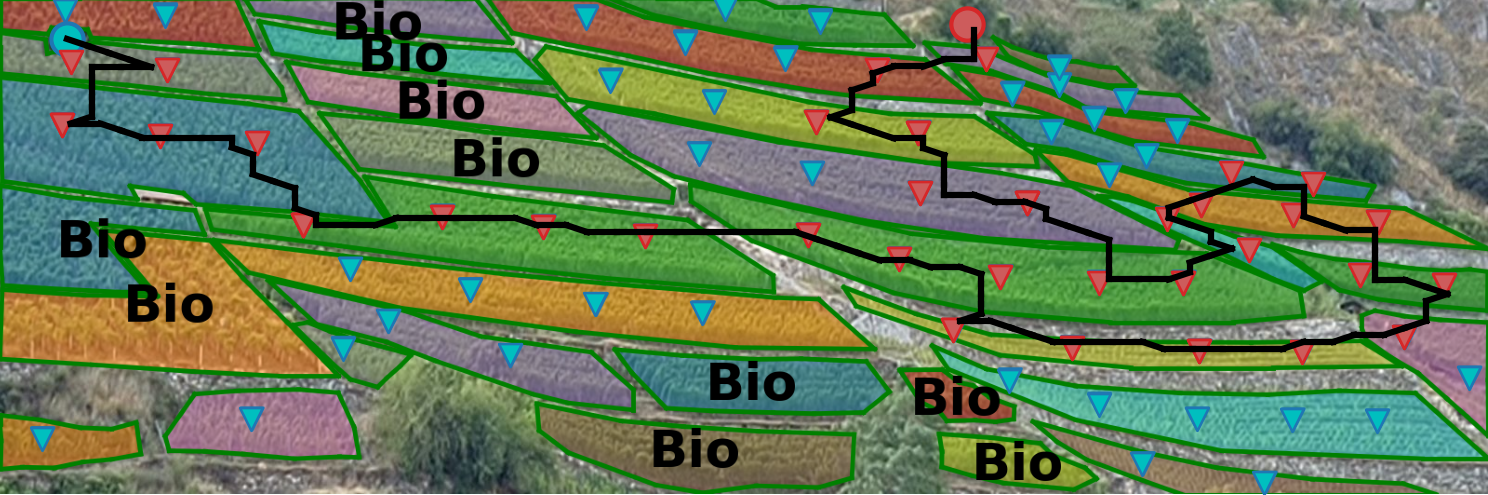}
      \caption{Best solution found.}
      \label{fig:solution}
    \end{subfigure}
\caption{A scenario with (a) cultivation and biocultivation areas, (b) their segmentation and spraying points (triangles), and (c) NaviFormer's solution from start to end depot (cyan and red circles).}
\label{fig:agriculture}
\end{figure}

\begin{figure}[t]
\centering
    \begin{subfigure}{0.325\columnwidth}
      \centering
      \includegraphics[width=\columnwidth]{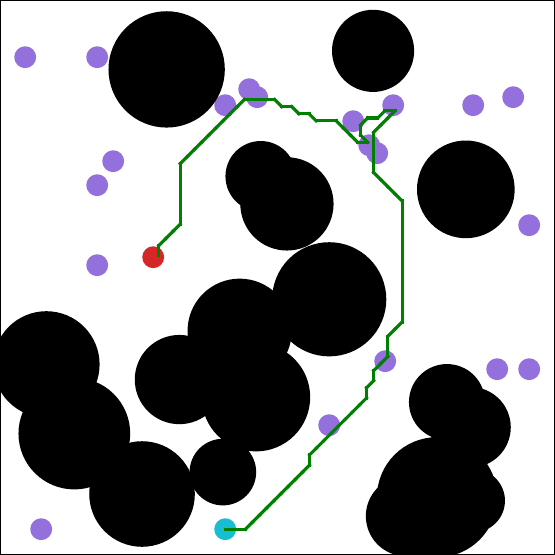}
      \caption{$n=20$.}
      \label{fig:synthetic20}
    \end{subfigure}
    \begin{subfigure}{0.325\columnwidth}
      \centering
      \includegraphics[width=\columnwidth]{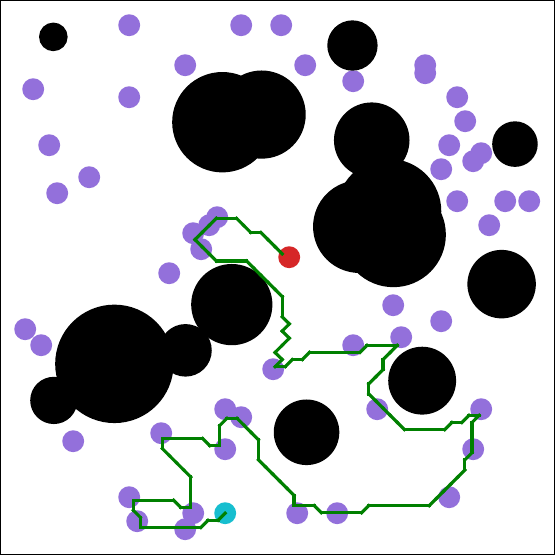}
      \caption{$n=50$.}
      \label{fig:synthetic50}
    \end{subfigure}
    \begin{subfigure}{0.325\columnwidth}
      \centering
      \includegraphics[width=\columnwidth]{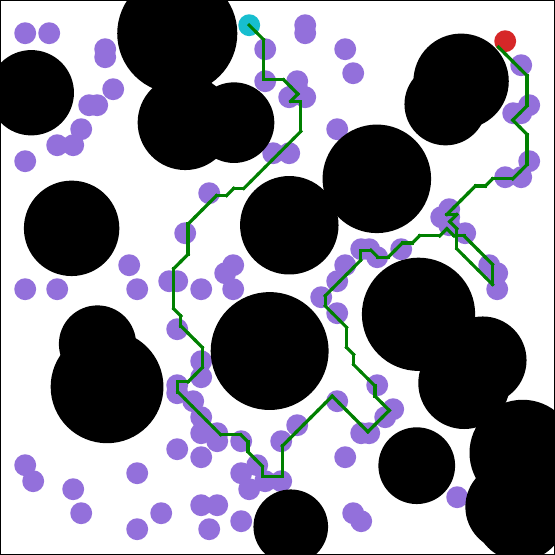}
      \caption{$n=100$.}
      \label{fig:synthetic100}
    \end{subfigure}
\caption{Solutions provided by NaviFormer for some random synthetic scenarios.}
\label{fig:synthetic}
\end{figure}

\begin{figure}[t]
\centering
    \begin{subfigure}{\columnwidth}
      \centering
      \includegraphics[width=\columnwidth]{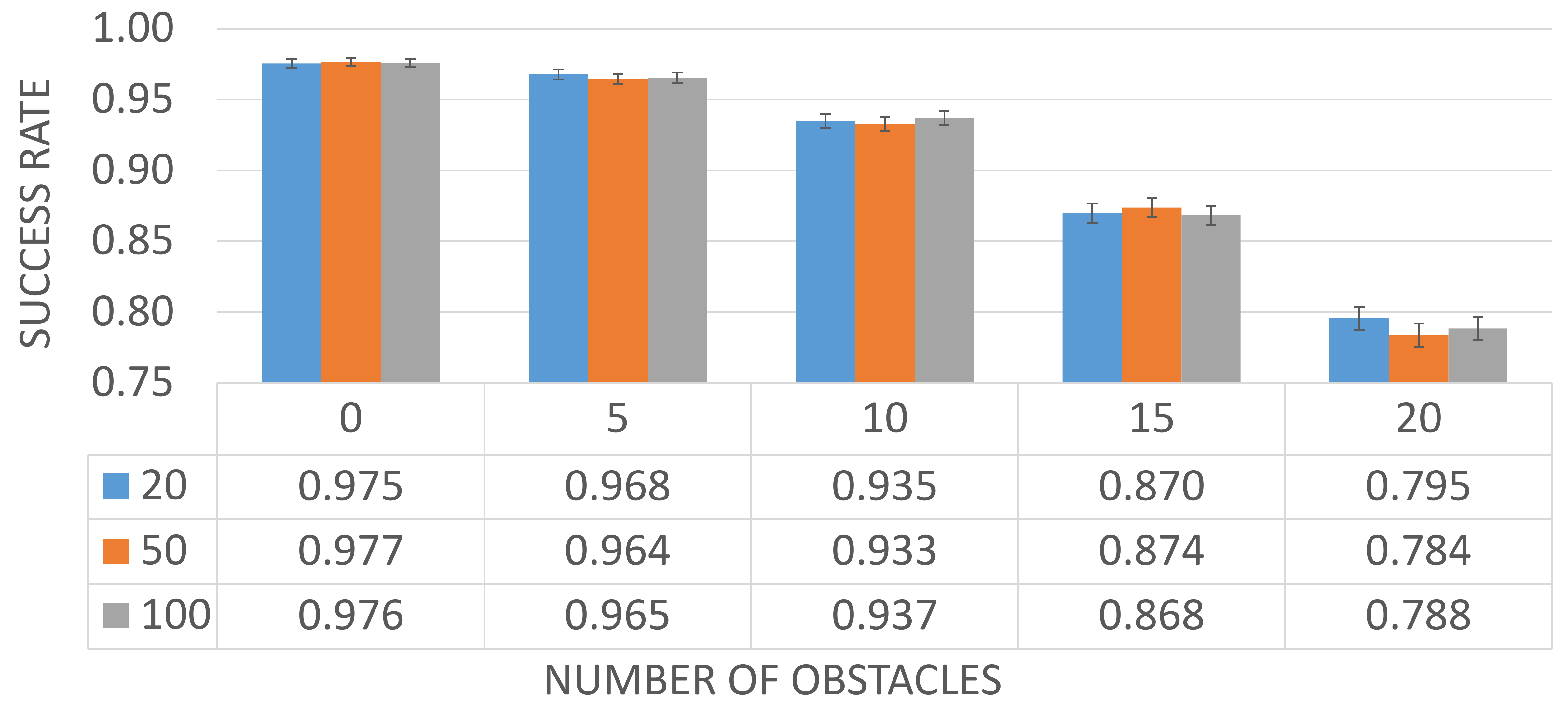}
      \caption{Success rate.}
      \label{fig:success}
    \end{subfigure}
    \begin{subfigure}{\columnwidth}
      \centering
      \includegraphics[width=\columnwidth]{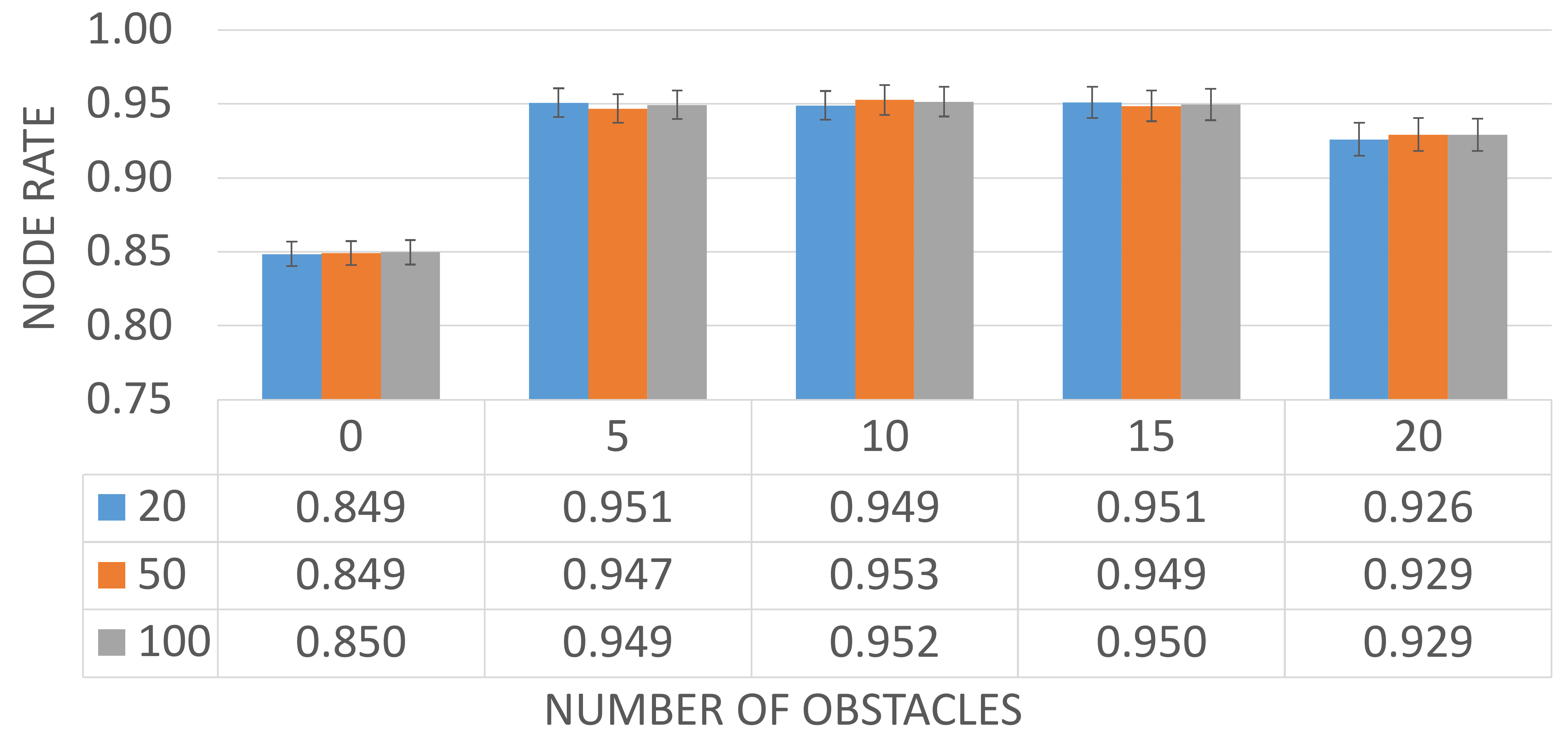}
      \caption{Node rate.}
      \label{fig:nodes}
    \end{subfigure}
\caption{System performance for small, medium, and large ($n=20, 50, 100$) synthetic scenarios.}
\label{fig:results}
\end{figure}



We evaluated NaviFormer\footnote{Code is publicly available at: \url{http://gti.ssr.upm.es/data}}
with real and synthetic data. Synthetic data comprises 1280k train samples, 10k validation samples, and 10k test samples, and covers small (20 nodes), medium (50 nodes), and large (100 nodes) scenarios. Node coordinates were sampled from the uniform distribution $\mathcal{U}(0, 1)$. A random number of obstacles (from 5 to 20) was sampled at random locations (also from $\mathcal{U}(0, 1)$) with radius $r^{obs} \sim \mathcal{U}(0.02, 0.12)$. The time limit $T$ was set to allow visiting around half of the nodes, since these cases tend to be more challenging \cite{Vansteenwegen2011}, resulting in $T=2, 3, 4$ for $n=20, 50, 100$. The time step was fixed at $t_s=0.02$.

Real data from a real-world application about pesticide spraying from an unmanned aerial vehicle (UAV) has also been used. This application includes two different types of areas to consider: cultivation and biocultivation (pesticide-free) areas (see Fig. \ref{fig:agriculture}). The UAV should cover the first ones with pesticide, and completely avoid flying over the latter ones. To fully cover each cultivation area, they are discretized into multiple spraying points, which are considered as nodes for the NOP. Moreover, the most restrictive constraint between the UAV's pesticide capacity and fuel is used for its arrival to the end depot. We adapted PASTIS dataset \cite{Garnot2021}, originally designed for image segmentation, for this task by normalizing the coordinates in the range [0, 1], setting $T$ similarly to the synthetic data, and randomly choosing biocultivation areas (ranging from 5 to 20) as obstacles (enclosed in circles like those from the synthetic data). The resulting number of PASTIS's instances is 2433, with an average of 30 nodes and 11 obstacles per instance.

To evaluate the performance of NaviFormer, we consider a couple of metrics: the success rate and the node rate. The success rate is defined as the rate of episodes where the agent reaches the depot on time without colliding over the total number of evaluated episodes. On the other hand, the node rate measures the number of nodes visited over the number of expected nodes to be visited. As it was commented before, the agent is expected to visit approximately half of the nodes. Notice that $T$ is fixed to approximately allow visiting half of the nodes, but an agent could visit slightly more nodes than the half, hence achieving node rates superior to 1.

Additional experiment-related information includes the use of the largest batch size that could fit on GPU (1024) to accelerate training and leverage a larger average loss function, which helps reduce variance. Other common hyperparameters, such as the network's hidden dimension, the number of encoding blocks ($N$), the learning rate, and the number of heads of the multi-head attention layers were set to the standard values of 128, 3, $1e^{-4}$, and 8, respectively. Also, we used the Adam optimizer. Finally, we performed all the experiments with 2 $\times$ Nvidia GeForce RTX 4090 (GPU) and a 13\textsuperscript{th} Gen. Intel Core i9-13900 $\times$ 32 (CPU).


\begin{table*}[t]
    \small 
    \caption{Ablation study showing the performance for synthetic scenarios with $n=50$ and $T=3$.}
    \label{tab:ablation}
    \begin{center}
        \begin{tabular}{l|rrrr}
            Ablation study (time in seconds) & Success rate & Node rate & Time (GPU) & Time (CPU) \\
            \hline
            \hline

            NaviFormer & \textbf{.904$\pm$.006} & .999$\pm$.003 & .280$\pm$.015 & .293$\pm$.012 \\

            NaviFormer (2-step) & .916$\pm$.006 & .876$\pm$.004 & .261$\pm$.019 & .274$\pm$.016 \\
            \hline

            w/ standard Transf. encoder & .886$\pm$.006 & \textbf{1.016$\pm$.003} & .277$\pm$.016 & .288$\pm$.013 \\

            w/ PN encoder-decoder & .814$\pm$.008 & .883$\pm$.003 & .332$\pm$.016 & .484$\pm$.012 \\
            
            w/ GPN encoder-decoder & .780$\pm$.008 & .893$\pm$.003 & .305$\pm$.020 & .461$\pm$.029 \\
            \hline
            
            


            w/ global maps & .838$\pm$.007 & .971$\pm$.035 & .290$\pm$.020 & .363$\pm$.016 \\

            w/o maps & .799$\pm$.008 & .876$\pm$.004 & \textbf{.130$\pm$.010} & \textbf{.191$\pm$.018} \\
            \hline

            w/ 4 directions & .866$\pm$.007 & .845$\pm$.003 & .279$\pm$.014 & .288$\pm$.011
        \end{tabular}
    \end{center}
\end{table*}



\begin{table*}[t]
    \small 
    \caption{Comparison with some baselines on the PASTIS and synthetic datasets.}
    \label{tab:comparison}
    \begin{center}
        \begin{tabular}{l|rr|rr|rr}
            \multirow{2}{*}{ALGORITHMS} & \multicolumn{2}{c|}{PASTIS DATASET} & \multicolumn{2}{c|}{SYNTHETIC DATASET} & \multicolumn{2}{c}{TIME (s)} \\
        
            & \multicolumn{1}{c}{Success rate} & \multicolumn{1}{c|}{Node rate} & \multicolumn{1}{c}{Success rate} & \multicolumn{1}{c|}{Node rate} & \multicolumn{1}{c}{GPU} & \multicolumn{1}{c}{CPU} \\
            \hline
            \hline

            NaviFormer & .891$\pm$.013 & \textbf{.905$\pm$.024} & .888$\pm$.006 & \textbf{.949$\pm$.010} & .280$\pm$.015 & .293$\pm$.012 \\

            NaviFormer + NA* & .822$\pm$.016 & .769$\pm$.021 & .856$\pm$.007 & .776$\pm$.008 & \textbf{.104$\pm$.014} & .212$\pm$.020 \\

            NaviFormer + A* & .932$\pm$.010 & .833$\pm$.021 & .933$\pm$.005 & .840$\pm$.009 & .107$\pm$.015 & .147$\pm$.013 \\

            NaviFormer + TransPath & .917$\pm$.011 & .832$\pm$.021 & .916$\pm$.006 & .837$\pm$.009 & .174$\pm$.011 & .369$\pm$.007 \\
            \hline

            PN + CNN & .870$\pm$.014 & .871$\pm$.021 & .862$\pm$.007 & .932$\pm$.010 & .336$\pm$.021 & .485$\pm$.031 \\

            PN + NA* & .842$\pm$.015 & .756$\pm$.018 & .897$\pm$.006 & .727$\pm$.007 & .109$\pm$.014 & .237$\pm$.023 \\

            PN + A* & .923$\pm$.011 & .815$\pm$.019 & .935$\pm$.005 & .783$\pm$.008 & .109$\pm$.012 & .164$\pm$.017 \\

            PN + TransPath & .896$\pm$.012 & .861$\pm$.021 & .914$\pm$.006 & .837$\pm$.008 & .177$\pm$.017 & .399$\pm$.007 \\
            \hline

            GPN + CNN & .832$\pm$.015 & .838$\pm$.023 & .783$\pm$.008 & .895$\pm$.011 & .318$\pm$.022 & .458$\pm$.032 \\

            GPN + NA* & .719$\pm$.018 & .699$\pm$.020 & .628$\pm$.010 & .710$\pm$.009 & .111$\pm$.167 & .228$\pm$.022 \\

            GPN + A* & .855$\pm$.014 & .791$\pm$.021 & .799$\pm$.008 & .790$\pm$.009 & .112$\pm$.012 & .178$\pm$.017 \\
            
            GPN + TransPath & .835$\pm$.015 & .789$\pm$.021 & .776$\pm$.008 & .787$\pm$.009 & .184$\pm$.024 & .377$\pm$.007 \\
            \hline
            
            OR-Tools + NA* & .799$\pm$.016 & .427$\pm$.009 & .862$\pm$.007 & .365$\pm$.003 & - & .139$\pm$.007 \\
            
            OR-Tools + A* & .874$\pm$.014 & .473$\pm$.009 & .909$\pm$.006 & .405$\pm$.004 & - & \textbf{.126$\pm$.006} \\
            
            OR-Tools + D* & .907$\pm$.012 & .554$\pm$.010 & .948$\pm$.004 & .451$\pm$.004 & - & 3.119$\pm$.068 \\
            \hline
            
            GA + NA* & .809$\pm$.016 & .662$\pm$.015 & .866$\pm$.007 & .627$\pm$.006 & - & 11.294$\pm$.758 \\
            
            GA + A* & .827$\pm$.016 & .746$\pm$.016 & .880$\pm$.007 & .716$\pm$.007 & - & 6.150$\pm$.279 \\
            
            GA + D* & \textbf{.936$\pm$.010} & \textbf{.905$\pm$.018} & \textbf{.973$\pm$.003} & .827$\pm$.007 & - & 16.034$\pm$1.373
            
        \end{tabular}
    \vspace{-0.5cm}
    \end{center}
\end{table*}

\textbf{Evaluation} NaviFormer's qualitative performance is illustrated in Fig. \ref{fig:solution} on real scenarios, which demonstrates its ability to visit 34 out of 71 spraying points. Moreover, Fig. \ref{fig:synthetic} illustrates the solutions found by NaviFormer across synthetic scenarios of varying sizes. The quantitative performance, instead, is depicted in Fig. \ref{fig:results} as a function of the number of nodes and obstacles for synthetic test scenarios. As expected, success rates decline with more obstacles, whereas node rates remain relatively static as the obstacle number grows. The first is intuitive, as obstacle-free scenarios are simpler. The latter is attributed to reduced node-to-node distances with abundant obstacles. Notice the anomaly for 0 obstacles, where node-to-node distances are higher, and it reduces the node rate. Besides, the system has not been trained for the case of 0 obstacles, since it is essentially the OP and could be better solved with straight paths. Comparing small, medium, and large scenarios, node rates are higher for larger ones due to the mentioned node-to-node distances. On the other hand, success rates decrease similarly for every number of nodes since it is independent of the success rate. Notice also that NaviFormer was trained for this experiment with scenarios of a variable number of regions, achieved by adding dummy nodes, to enhance its flexibility and generalization capability.

\textbf{Ablation Study} We carried out four ablation studies in Tab. \ref{tab:ablation} by removing and/or substituting different modules of the network and evaluating the resulting models on medium-sized synthetic scenarios ($n=50$, $T=3$). First, we conducted an experiment to confirm that the proposed joint NaviFormer network (simultaneously acts as route and path planner) performs better than training both components in isolation. We pretrained the base Transformer network (route planner) included in NaviFormer to predict routes for the OP. Later, this model was employed to fit the lightweight CNN (path planner) for the navigation task. The results confirm a deterioration in node rate, which suggests that the behavior of the route planner is influenced by the path planner and vice versa. Besides, the 1-step approach is end-to-end trainable. Secondly, the combined attention encoder is assessed by evaluating the model with a traditional Transformer encoder, such as those from \cite{Kool2019,Fuertes2022}, that receives the nodes and obstacles separately and combines them at the state embedding. This approach increases the node rate at the expense of the success rate, adopting a more risk-taking policy that could be detrimental in scenarios with a higher obstacle density. Besides, we replaced the entire Transformer architecture with two RNN-based alternatives (PN and GPN), obtaining lower results with slightly larger computation times. This demonstrates the superiority of Transfomers over RNNs. Third, the importance of local maps to predict the direction is analyzed. The local maps are limited to represent the surroundings of the agent, while the assessed global maps contain the entire scenario. The inclusion of global maps provides lower performance due to the larger compression of the map by the convolutional layers, and increases the computational cost. On the other hand, the removal of maps (NaviFormer considers only linear projections of obstacles and start/goal coordinates for each time step) deteriorates success rate in exchange of faster performance. Fourth and lastly, we observed that reducing the number of possible directions decreases node rate, due to limited mobility.



\textbf{Comparison} NaviFormer is compared to 2-step baselines on both PASTIS and synthetic datasets, including combinations of A* \cite{Mandloi2021}, Neural A* (NA*) \cite{Yonetani2021}, D* \cite{Ravankar2017}, TransPath \cite{Kirilenko2023}, and NaviFormer's CNN with PN \cite{Gama2021}, Graph PN (GPN) \cite{Ma2020}, OR-Tools \cite{Ortools2023}, GA \cite{Xiao2022}, and NaviFormer's Transformer. To compensate the suboptimal performance of obstacle-agnostic route planners (OR-Tools and GA), we allow these 2-step methods to reduce $T$ by a small $\epsilon=0.1$ for \{OR-Tools, GA\}+D*, and $\epsilon=0.3$ for \{OR-Tools, GA\}+\{A*, NA*\}, which facilitates arriving at the end depot on time, but can negatively affect the node rate. Tab. \ref{tab:comparison} highlights that NaviFormer's performance achieves the highest node rate, proving the importance of tackling the problem holistically. Only GA + D* effectively competes with NaviFormer in terms of node rate, although its computational cost is comparatively huge (several orders of magnitude slower). Regarding success rate, NaviFormer is slightly surpassed by other path planning-designed algorithms that ensure obstacle avoidance but can fail in reaching the end depot on time (despite of the inclusion of the $\epsilon$ term). Our proposal still provides a very competitive success rate and performs as fast as most of the presented methods, which is crucial for real-time applications. In fact, the fastest proposal is NaviFormer + \{NA*, A*\}, which could be considered for applications requiring extremely low computation times. Moreover, NaviFormer reaches the best balance between success and node rates, proving the importance of solving the problem holistically.

\section{Conclusions}
\label{sec:conclusions}

We proposed a novel DRL approach, called NaviFormer, to solve the holistic navigation problem. NaviFormer combines route planning (waypoint sequences) and path planning (collision-free trajectories) using a novel Transformer-based encoder to efficiently create joint embeddings for waypoints and obstacles, allowing the prediction of next waypoints to visit and safe directions to reach them. Compared to 2-step state-of-the-art methods, NaviFormer achieves a desirable balance between success rate (ability to reach the end depot on time without colliding), node rate (number of nodes visited per episode) and computation time, making it suitable for real-time applications. Future research could focus on improving direction prediction for a wider range of motion actions, possibly through continuous action spaces for smoother trajectories. In addition, we observed scalability issues as the number of obstacles grows, mainly due to not hard-imposing their avoidance like other state-of-the-art works. Moreover, extending the approach to support irregularly shaped obstacles could improve generalization. Finally, dynamic environment handling could also be explored since holistic solutions could be very beneficial in these cases.


\bibliographystyle{IEEEtran}
\bibliography{refs}

\end{document}